\documentclass{article}
\usepackage{amsmath}
\usepackage{graphicx}
\usepackage{booktabs}
\usepackage{multirow}
\usepackage[most]{tcolorbox}
\usepackage{graphicx}
\usepackage{caption}
\usepackage{url}
\usepackage{makecell}
\usepackage{tcolorbox}
\usepackage{array}

\usepackage{booktabs}
\usepackage{graphicx}

\usepackage{tabularx}
\usepackage{array}
\usepackage{tikz}      
\usepackage{framed}    
\usepackage{enumitem}  
\usepackage{xcolor}    
\tcbuselibrary{skins}
\usepackage[final]{corl_2026} 

\title{PathPainter: Transferring the Generalization Ability of Image Generation Models to Embodied Navigation}

\author{
  \textbf{Yijin Wang}$^{1,2,*}$ \quad
  \textbf{Yuru Tian}$^{1,2,*}$ \quad
  \textbf{Xijie Huang}$^{1,2,*}$ \quad
  \textbf{Weiqi Gai}$^{2,3}$ \\
  \textbf{Mo Zhu}$^{1,2}$ \quad
  \textbf{Xin Zhou}$^{2}$ \quad
  \textbf{Yuze Wu}$^{1,2,\dagger}$ \quad
  \textbf{Fei Gao}$^{1,2,\dagger}$
}

\begin{document}
\maketitle

\begingroup
\renewcommand{\thefootnote}{}
\footnotetext{
$^{1}$Zhejiang University.
$^{2}$Differential Robotics.
$^{3}$Beihang University.
$^{*}$Equal contribution.
$^{\dagger}$Corresponding authors: 
\texttt{wuyuze000@zju.edu.cn}, \texttt{fgaoaa@zju.edu.cn}.
}
\endgroup


\begin{abstract}
Bird's-eye-view (BEV) images have been demonstrated to provide valuable prior information for navigation. Given the global information provided by such views, two key challenges remain: how to fully exploit this information and how to reliably use it during execution. In this paper, we propose a navigation system that uses BEV images as global priors and is designed for ground and near-ground robotic platforms. The system employs an image generation model to interpret human intent from natural language, identify the target destination, and generate traversability masks. During execution, we introduce cross-view localization to align the robot's odometry with the BEV map and mitigate long-term drift in conventional odometry. We conduct extensive benchmark experiments to evaluate the proposed method and further validate its effectiveness on both UAV and quadruped robot platforms. Using only a conventional local motion planner, the quadruped robot successfully completes a 285-meter outdoor long-range navigation task. This work demonstrates how the world-understanding capabilities of foundation models can be transferred to embodied navigation, enabling robots to benefit from the strong generalization ability of existing image generation models.




\end{abstract}

\keywords{Robot Navigation, BEV Prior, Foundation Models, Cross-view Localization, Image Generation Model}

\section{Introduction}
BEV images from aerial or satellite perspectives provide robots with valuable global priors for outdoor navigation, including road topology, open areas, obstacle layouts, and spatial relationships between targets and surrounding structures~\cite{elnoor2024robot,klammer2024bevloc,zhang2025dual}. These priors are particularly important for natural-language-guided navigation, where robots must ground language instructions in complex scenes and infer feasible routes~\cite{lee2025citynav,huang2023visual}. As a result, bird's-eye-view information has been widely used in air-ground collaborative navigation~\cite{KumarCollabration,MojuZhaoCollabration,li2024colag,deng2025tightly}, satellite-map-based navigation~\cite{elnoor2024robot,klammer2024bevloc,huang2023stochastic}, and high-low altitude collaborative navigation~\cite{wu2025aeroduo,munasinghe2024comprehensive,zhang2024air}.

Existing BEV-based navigation systems still struggle to convert aerial
observations into executable navigation priors. Many methods compress BEV images
into semantic graphs, topological maps, or category-level representations for
efficient planning~\cite{KumarCollabration,MojuZhaoCollabration}. Although
compact, these representations may discard geometric details and visual cues
needed for navigation, such as road boundaries, passage width, obstacle layout,
and open-space continuity.

Moreover, traversability, i.e., whether a region can support feasible robot
movement, is often approximated by human-predefined semantic classes,
especially roads~\cite{KumarCollabration}. Such road-centric priors are effective in structured scenes, but fail to capture diverse traversable regions in open outdoor environments, where feasible navigation may involve sidewalks, trails, off-road routes, and movement across plazas, beaches, and other open spaces.
\begin{figure}[!t]
\centering
\includegraphics[width=1\linewidth]{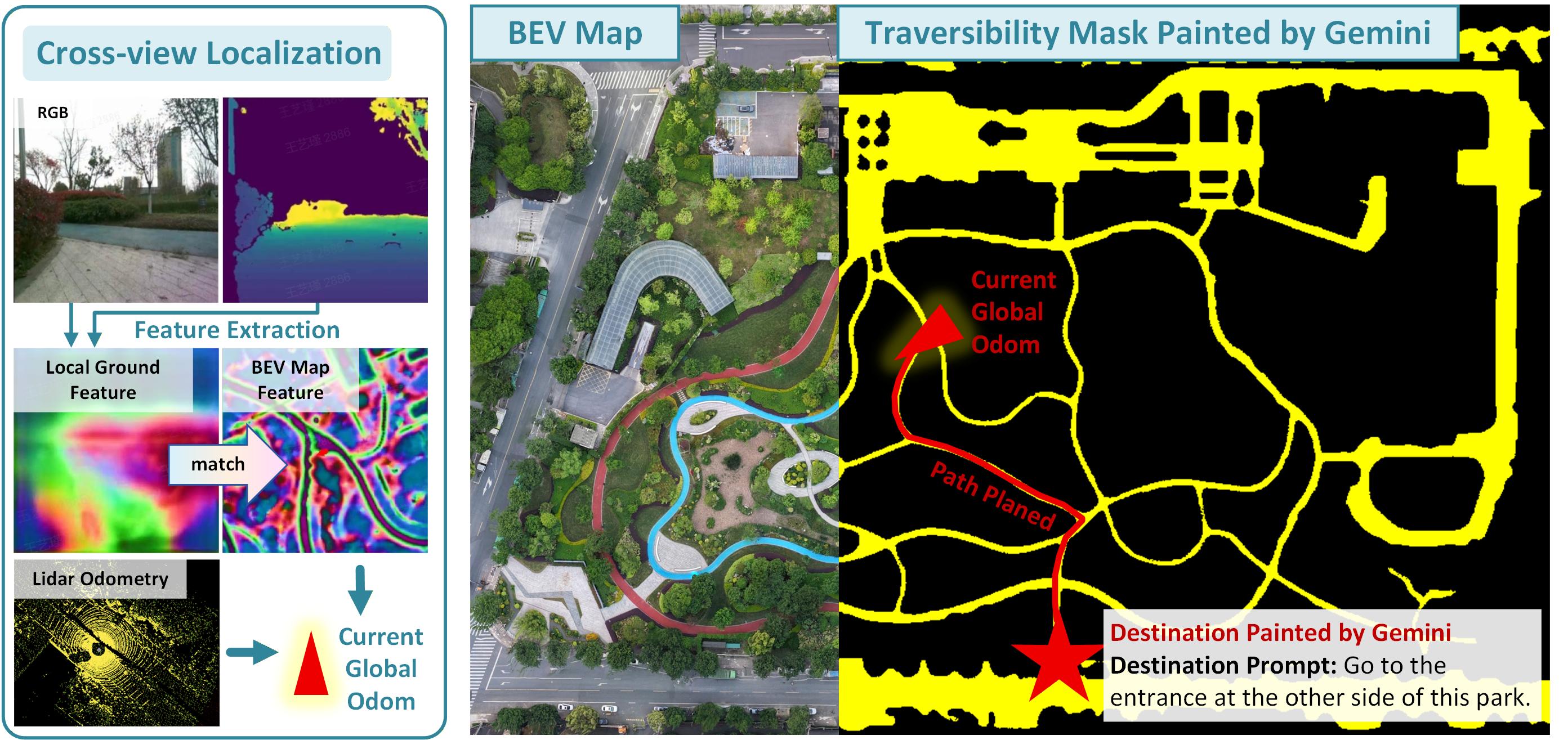}
\caption{\textbf{Overview of the Navigation System.}
\textbf{Left:} Cross-view localization extracts embeddings from local ground features reconstructed from RGB-D observations and matches them with feature embeddings from the BEV map to estimate the robot's global odometry.
\textbf{Right:} Given the destination prompt and the BEV map, the image generation model marks the target region with a generated star marker and produces a traversability mask. The red path is obtained by running $\mathrm{A}^*$ on the mask.}

\label{fig:head}
\end{figure}


Recent foundation models have demonstrated strong generalization capabilities in visual, language, and spatial understanding. In particular, image generation models~\cite{google2025nanobananapro,gptimage2} encode rich priors about spatial layout, object relations, and navigability cues. Prior work has shown that visual understanding tasks can be reformulated as image generation problems~\cite{gabeur2026imagegeneratorsgeneralistvision}.

These observations motivate our central question: can BEV navigation be
reformulated as an image generation problem, allowing image generation models to
produce dense navigation priors for target grounding, traversability estimation,
and global path planning, and further convert them into paths that are
structured, searchable, and executable by real robots?

In this paper, we propose PathPainter, a path-planning framework for natural-language-conditioned BEV navigation, as shown in Fig. \ref{fig:head}. Given a BEV image, the robot's start position, and a language instruction, PathPainter infers the target region, generates a navigation-oriented traversability mask, and applies $\mathrm{A}^*$\cite{A*} as the trajectory planner to obtain an executable global path. We further integrate PathPainter into a real-world long-range navigation system that combines cross-view localization, BEV-based global planning, and local motion execution. Cross-view localization aligns the robot trajectory with the BEV map without relying on precise GPS/RTK localization.

PathPainter demonstrates a practical way to transfer the generalization ability of image generation models to embodied navigation through BEV prior interpretation. Our main contributions are:
(1) We propose PathPainter, which transfers generative vision priors to natural-language-conditioned BEV navigation through destination inference, traversability-mask generation, and search-based planning.
(2) We develop a real-world long-range navigation system combining cross-view localization, BEV global planning, and local motion planning for near-ground outdoor robots.
(3) We validate PathPainter on real-world navigation and path-generation benchmarks, demonstrating its robustness. Code and data will be released upon publication.

\section{Related Work}
\subsection{Navigation with Priors}

Compared with onboard local observations, prior information provides larger-scale environmental structure and potential goal locations, reducing exploration cost in long-range navigation. Existing works use priors such as OpenStreetMap~\cite{lee2025citynav,li2025urbanvla}, abstract topological maps~\cite{tan2025mobile}, and aerial maps~\cite{moore2023ura,shair2008use} for global planning, subgoal selection, or local planning constraints. To use these priors for navigation, prior works often abstract them into road graphs, topological nodes, or semantic regions, and study how to align onboard observations with the global prior during execution. While useful for planning and localization, such abstractions may discard fine-grained geometry, visual appearance, and local traversability cues in aerial images. In contrast, our work directly extracts traversable regions from aerial images for $\mathrm{A}^*$-based planning on the resulting masks, and introduces cross-view localization to align robot odometry with the aerial prior during execution.

%
%

\subsection{Navigation with Foundation Models}

Recent works have introduced foundation models into navigation, enabling robots to understand human instructions and perform goal, object, and language-guided navigation in complex environments. These methods mainly follow two paradigms: vision-language models directly predict actions or high-level waypoints from instructions~\cite{zhang2024navid,wu2025vla}, video generation models support executable action inference through instruction-conditioned future prediction~\cite{zhang2026sparse,huang2026navdreamer,huang2026navgen,zhu2026diffwam}. Some methods further combine both paradigms~\cite{hu2025astranav}. Overall, these approaches demonstrate the semantic understanding and cross-scene generalization of foundation models, advancing open-world navigation.

\section{Method}
\subsection{Pipeline}

\begin{figure}[!b]
    \centering
    \includegraphics[width=1.0\linewidth]{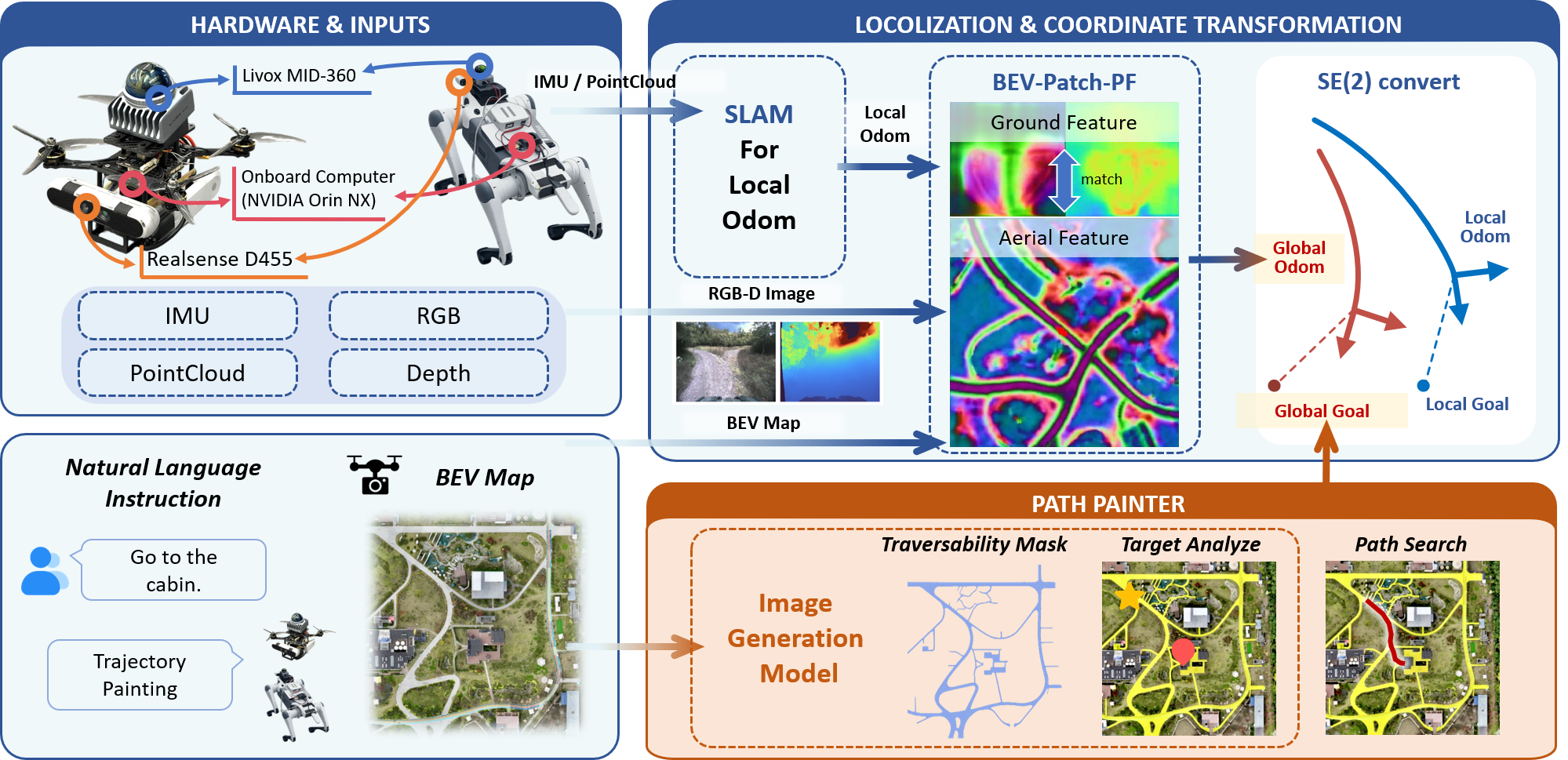}
    \caption{Pipeline of our navigation system.}
    \label{fig:pipeline}
\end{figure}

As shown in Fig.~\ref{fig:pipeline}, this paper proposes a bird's-eye-view-based navigation planning framework. The system takes an aerial orthophoto and a natural-language instruction as input, outputs an executable navigation path, and deploys it on a real robotic platform for autonomous execution.

The overall pipeline consists of two levels: high-level path generation and low-level motion execution. At the high level, the system first uses an image generation model to semantically understand the BEV map, draw the traversability mask, and extract a global path based on the start and goal positions. At the low level, the system maps the global path to the robot's local coordinate frame through cross-view localization, and combines LiDAR odometry with a robust local planner~\cite{zhou2020ego} for real-time obstacle avoidance and trajectory tracking.

This hierarchical design decouples high-level semantic reasoning from low-level motion control, enabling stable navigation in complex and unstructured outdoor environments.

\begin{figure}[!t]
    \centering
    \includegraphics[width=1.0\linewidth]{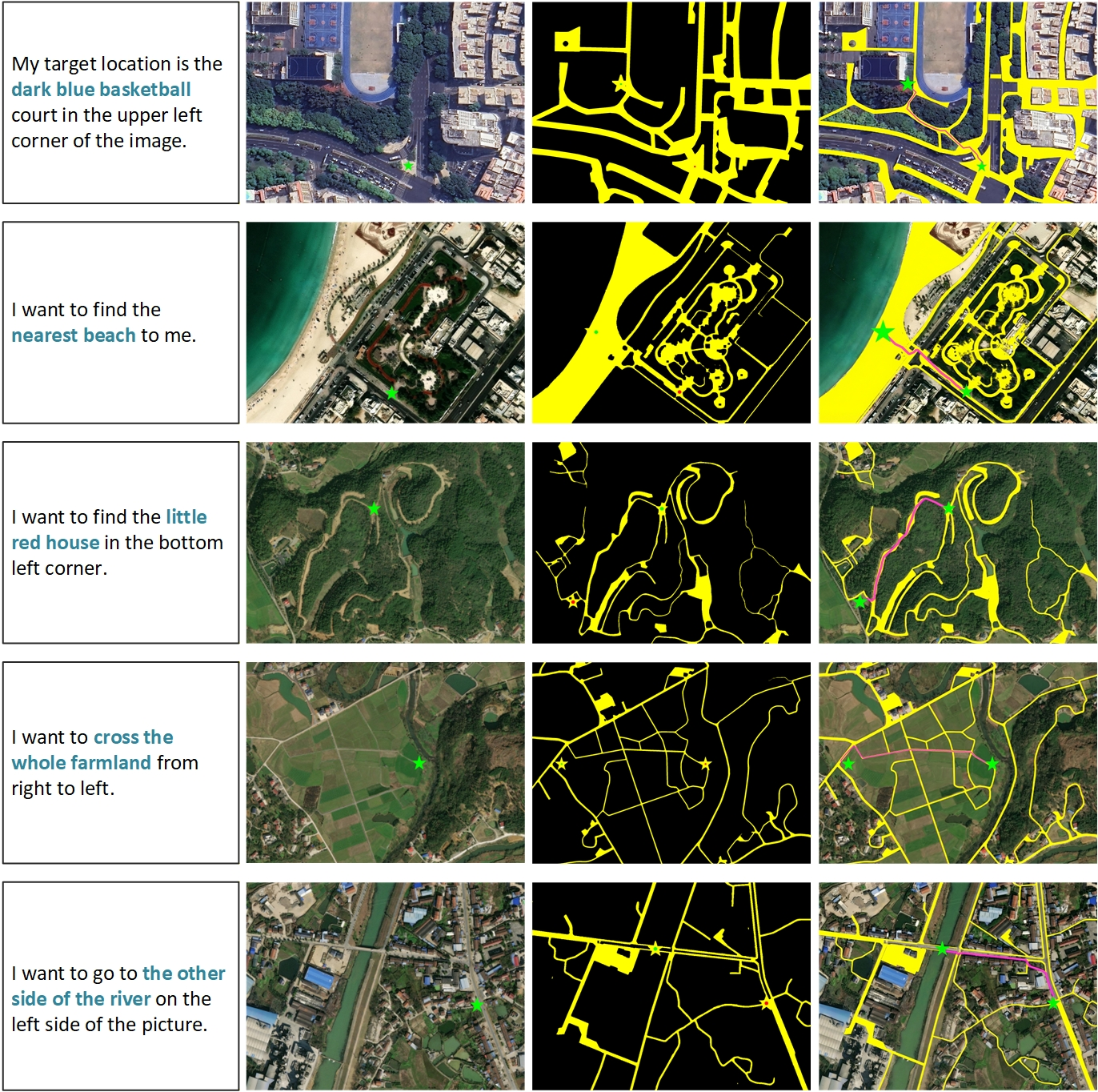}
    \caption{
\textbf{Workflow of PathPainter. }
\textbf{Column 1:} natural-language destination query. 
\textbf{Column 2:} original map with the current robot position. 
\textbf{Column 3:} Traversability mask.
\textbf{Column 4:} final planning result, where the generated traversability mask, predicted goal position, and planned path are overlaid on the original map.
}
    \label{fig:mask}
\end{figure}

\subsection{Extracting Executable Paths via Image-to-Image Generation}



We formulate the path planning problem as an image-to-image generation process, thereby leveraging the spatial understanding and semantic reasoning capabilities of image generation models to transform high-level map understanding into an executable navigation path. As shown in Fig.~\ref{fig:mask}, the image generation model is responsible for two main tasks: goal position prediction and traversability-mask segmentation. Specifically, we first mark the robot's current position on the BEV map with a green star. Conditioned on this starting point and the natural-language instruction, the image generation model is prompted to infer and draw the target position.

Meanwhile, the system employs carefully designed prompts to guide the image generation model to generate an image mask representing the traversable area, referred to as the traversability mask. This mask explicitly encodes walkable regions in the scene, such as roads, sidewalks, and open spaces, while suppressing non-traversable areas such as buildings, vegetation, and obstacles. For road regions that are partially occluded by trees but remain semantically continuous, the model can preserve their connectivity based on global contextual reasoning. This process can therefore be regarded as a mapping from raw visual input to a navigable-space representation.

Before path extraction, we apply a hallucination detection mechanism to the generated output, as illustrated in Fig.~\ref{huanjue}.  Since the prompt in Appendix~A.1 requires masks to be overlaid on the original image, we compare corresponding non-mask pixels in the output and input images. A mean Euclidean RGB distance $\geq 50$ triggers regeneration. On the evaluated benchmark, detection was 100\% accurate, and all 14 severe hallucination cases were corrected within two regeneration attempts. All Gemini metrics reported for this benchmark were computed using the corrected masks.

Finally, we extract a feasible start-to-goal path from the resulting traversability mask using $\mathrm{A}^*$ search. To improve safety, we add a boundary-distance penalty that favors path points farther from the traversable-region boundary, encouraging the path to stay near the road center.

\begin{figure}
    \centering
    \includegraphics[width=1\linewidth]{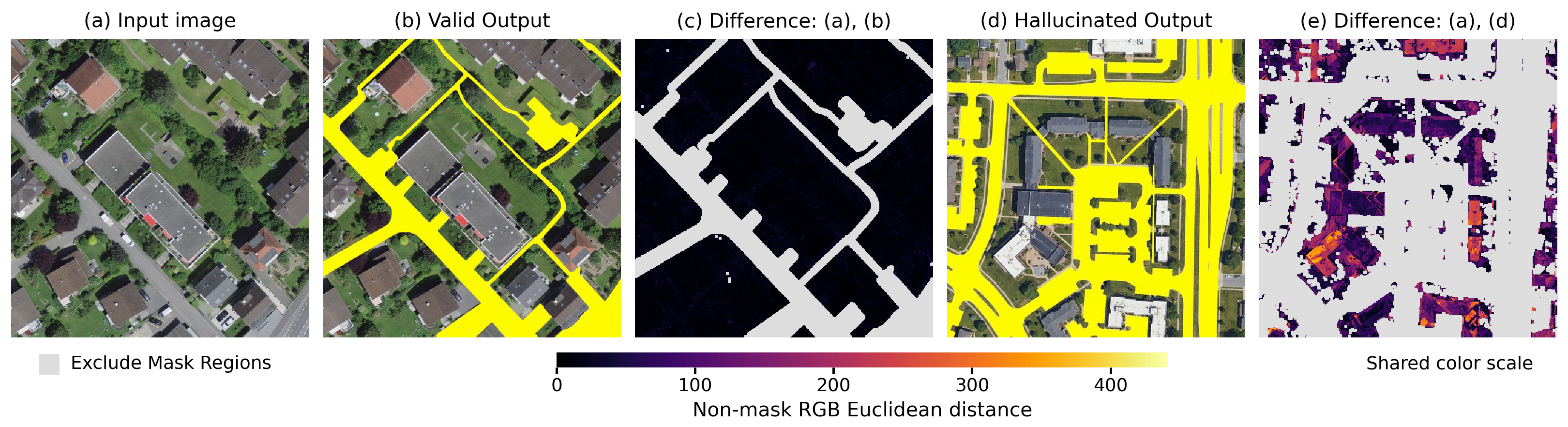}
\caption{\textbf{Hallucination Detection.}
\textbf{(a)} Input BEV image.
\textbf{(b)} Valid mask-overlay output.
\textbf{(c)} Pixel differences between (a) and (b).
\textbf{(d)} Hallucinated output for the same input.
\textbf{(e)} Pixel differences between (a) and (d).
Both difference maps visualize pixelwise Euclidean RGB distances outside the mask regions using a shared color scale; gray indicates excluded regions.}
        \label{huanjue}
\end{figure}

\subsection{Cross-view Localization}

Local odometry methods, such as FAST-LIO2\cite{xu2022fast}, provide high-rate motion estimates but do not directly recover the robot's global pose and may suffer from long-term drift. Therefore, executing a BEV global path requires continuous global localization to keep the local odometry frame aligned with the BEV map. Here, we use BEV-Patch-PF~\cite{lee2025bev}, which estimates the robot pose on the BEV map as $\mathbf{x}_t^M = (x_t^M, y_t^M, \theta_t^M)$, where $M$ denotes the map frame. Given the global path $\mathcal{P}^M = \{\mathbf{p}_1^M, \ldots, \mathbf{p}_K^M\}$, we select a look-ahead waypoint $\mathbf{p}_k^M$ according to the current global odometry and transform it into the local odometry frame $O$:
\begin{equation}
    \tilde{\mathbf{p}}_k^O = \mathbf{T}_{M}^{O} \tilde{\mathbf{p}}_k^M,
\end{equation}
where $\mathbf{T}_{M}^{O}$ is computed from cross-view localization and current odometry, and $\tilde{\mathbf{p}}$ denotes the homogeneous waypoint coordinate. The transformed waypoint is sent to the local planner as a short-horizon goal, enabling high-rate local planning with periodic drift correction.

\section{Experiment}
\subsection{Evaluating Path Generation Methods}

  It is important to note that our goal is not pure semantic segmentation or
complete road-topology reconstruction. Instead, we target \textbf{traversability-prior generation for navigation}. This task requires (1) recognizing visible
roads, (2) reasoning about continuity under occlusion and plausible
traversability in ambiguous regions, and (3) capturing diverse forms of traversability beyond road-like structures.
Although segmentation models, including SAM 3.1~\cite{carion2025sam3segmentconcepts} and Text2Seg~\cite{zhang2023text2seg}, as well as road-topology extraction models, including RNGDet++~\cite{xu2023rngdet++} and SAMRoad~\cite{hetang2024segment}, are strong within their own design objectives, their outputs are not necessarily optimal for downstream path planning. Nevertheless, we benchmark these representative methods from two perspectives:
their basic ability to produce stable visible-road priors and their utility for
downstream path planning.

\begin{figure}[!t]
  \centering
  \scriptsize
  \renewcommand{\arraystretch}{1.05} 
  \begin{minipage}[b]{0.5\textwidth}
    \centering
    \setlength{\tabcolsep}{4pt} 
\begin{tabular}{llccccc}
\toprule
Dataset & Method & IoU & Prec. & Rec. & F1 & Time (s) \\
\midrule
\multirow{4}{*}{\makecell{DeepGlobe\\\cite{demir2018deepglobe}}}
& GPT~\cite{gptimage2} & \underline{0.373} & 0.415 & \textbf{0.788} & \underline{0.544} & 70.69 \\
& Gemini~\cite{google2025nanobananapro} & \textbf{0.484} & \underline{0.648} & \underline{0.657} & \textbf{0.652} & 30.10 \\
& SAM 3.1~\cite{carion2025sam3segmentconcepts} & 0.367 & \textbf{0.651} & 0.456 & 0.536 & \textbf{0.17} \\
& Text2Seg~\cite{zhang2023text2seg} & 0.040 & 0.041 & 0.600 & 0.077 & \underline{0.46} \\
\midrule
\multirow{4}{*}{UAVid~\cite{lyu2020uavid}}
& GPT~\cite{gptimage2} & 0.501 & 0.528 & \textbf{0.906} & 0.667 & 71.47 \\
& Gemini~\cite{google2025nanobananapro} & \underline{0.698} & \underline{0.811} & \underline{0.834} & \underline{0.822} & 31.09 \\
& SAM 3.1~\cite{carion2025sam3segmentconcepts} & \textbf{0.698} & \textbf{0.885} & 0.768 & \textbf{0.822} & \textbf{0.20} \\
& Text2Seg~\cite{zhang2023text2seg} & 0.363 & 0.501 & 0.569 & 0.533 & \underline{0.51} \\
\midrule
\multirow{4}{*}{VDD~\cite{yeshchenko2020vdd}}
& GPT~\cite{gptimage2} & 0.392 & 0.411 & \textbf{0.894} & 0.563 & 69.48 \\
& Gemini~\cite{google2025nanobananapro} & \underline{0.607} & \underline{0.659} & \underline{0.884} & \underline{0.755} & 33.47 \\
& SAM 3.1~\cite{carion2025sam3segmentconcepts} & \textbf{0.666} & \textbf{0.787} & 0.813 & \textbf{0.800} & \textbf{0.17} \\
& Text2Seg~\cite{zhang2023text2seg} & 0.210 & 0.274 & 0.473 & 0.347 & \underline{0.46} \\
\bottomrule
\end{tabular}
    \captionof{table}{\textbf{Visible-road segmentation benchmark. }All methods segment only the road category.}
    \vspace{0mm} 
    \label{tab:path_generation_eval}
  \end{minipage}
  \hfill
  \begin{minipage}[b]{0.45\textwidth}
    \centering
    \includegraphics[width=\linewidth]{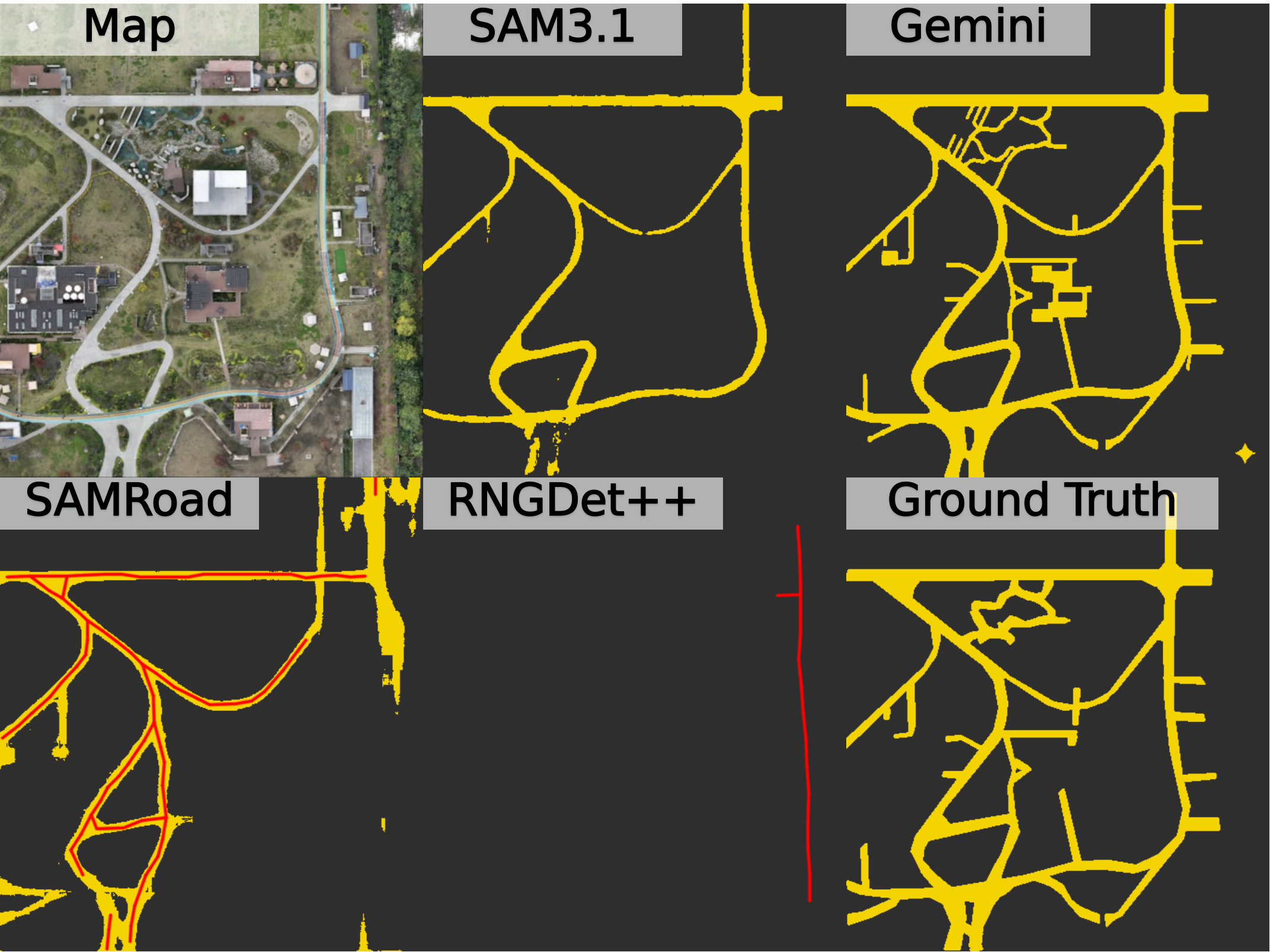}
    \vspace{-3mm} 
    \captionof{figure}{Real-world test on highly out-of-distribution scenes.}
    \label{fig:ooood}
  \end{minipage}
\end{figure}

\paragraph{Visible-road segmentation benchmark.}
We first compare our generation-based approach with segmentation models in terms of zero-shot visible-road segmentation ability on three UAV/satellite datasets: DeepGlobe~\cite{demir2018deepglobe}, UAVid~\cite{lyu2020uavid}, and VDD~\cite{yeshchenko2020vdd}. For each dataset, we
  use a fixed random split for evaluation. All methods are tested without
  task-specific training, fine-tuning, or manual correction. For both image
  generation models and open-vocabulary segmentation baselines, we use the same
  class prompt \verb|road|. Generated segmentation maps are converted into binary
  road masks before evaluation.
  Table~\ref{tab:path_generation_eval} shows differences between
  generation-based models and segmentation baselines. GPT and Gemini generally
  achieve substantially higher recall, indicating that generative models miss fewer ground-truth road pixels.
  In contrast, SAM 3.1 achieves higher
  precision and much lower inference time, making it a more conservative and
  reliable visible-road segmenter. Among the generation-based methods, Gemini provides the best balance between precision and recall, so we use it as the generation-based road-mask provider in the downstream task.


 \vspace{-3mm} 
\begin{figure}[!t]
  \centering
  \scriptsize
  \renewcommand{\arraystretch}{1.05} 
  \begin{minipage}[b]{0.6\textwidth}
    \centering
    \setlength{\tabcolsep}{4pt} 
\begin{tabular}{lccccccc}
\toprule
\multirow{2}{*}{Method} &
\multicolumn{3}{c}{In-domain (CityScale)~\cite{he2020sat2graph}} &
\multicolumn{3}{c}{OOD (Global-Scale)~\cite{yin2024satelliteimageroadgraph}} &
\multirow{2}{*}{Time (s) } \\
\cmidrule(lr){2-4}\cmidrule(lr){5-7}
& Succ.  & Valid.  & Len. 
& Succ.  & Valid.  & Len. & \\
\midrule
Gemini~\cite{google2025nanobananapro}
& \underline{0.902} & 0.932 & \underline{1.007}
& \textbf{0.766} & 0.904 & \underline{1.071} & 61.60 \\
Gemini-Direct~\cite{google2025nanobananapro}
& 0.280 & 0.910 & 1.242
& \underline{0.293} & 0.828 & \textbf{1.038} & 86.58 \\
SAMRoad~\cite{hetang2024segment}
& 0.853 & \textbf{0.994} & 1.095
& 0.339 & \textbf{0.975} & 1.164 & \textbf{9.34} \\
RNGDet++~\cite{xu2023rngdet++}
& \textbf{0.972} & \underline{0.969} & \textbf{1.005}
& 0.252 & \underline{0.949} & 1.148 & \underline{39.95} \\
SAM 3.1~\cite{carion2025sam3segmentconcepts}
& 0.018 & 0.016 & 0.829
& 0.042 & 0.040 & 0.887 & 0.266 \\
\bottomrule
\end{tabular} 
    \captionof{table}{\textbf{Downstream path-planning benchmark.} All methods are evaluated only on the road category. Valid. (Validity) and Len. (Length Ratio) are computed only over successful samples. The Len. metric indicates better performance when its value is closer to 1.}
    \vspace{0mm}
    \label{tab:downstream_path_eval}
  \end{minipage}
  \hfill
  \begin{minipage}[b]{0.32\textwidth}
    \centering
    \includegraphics[width=\linewidth]{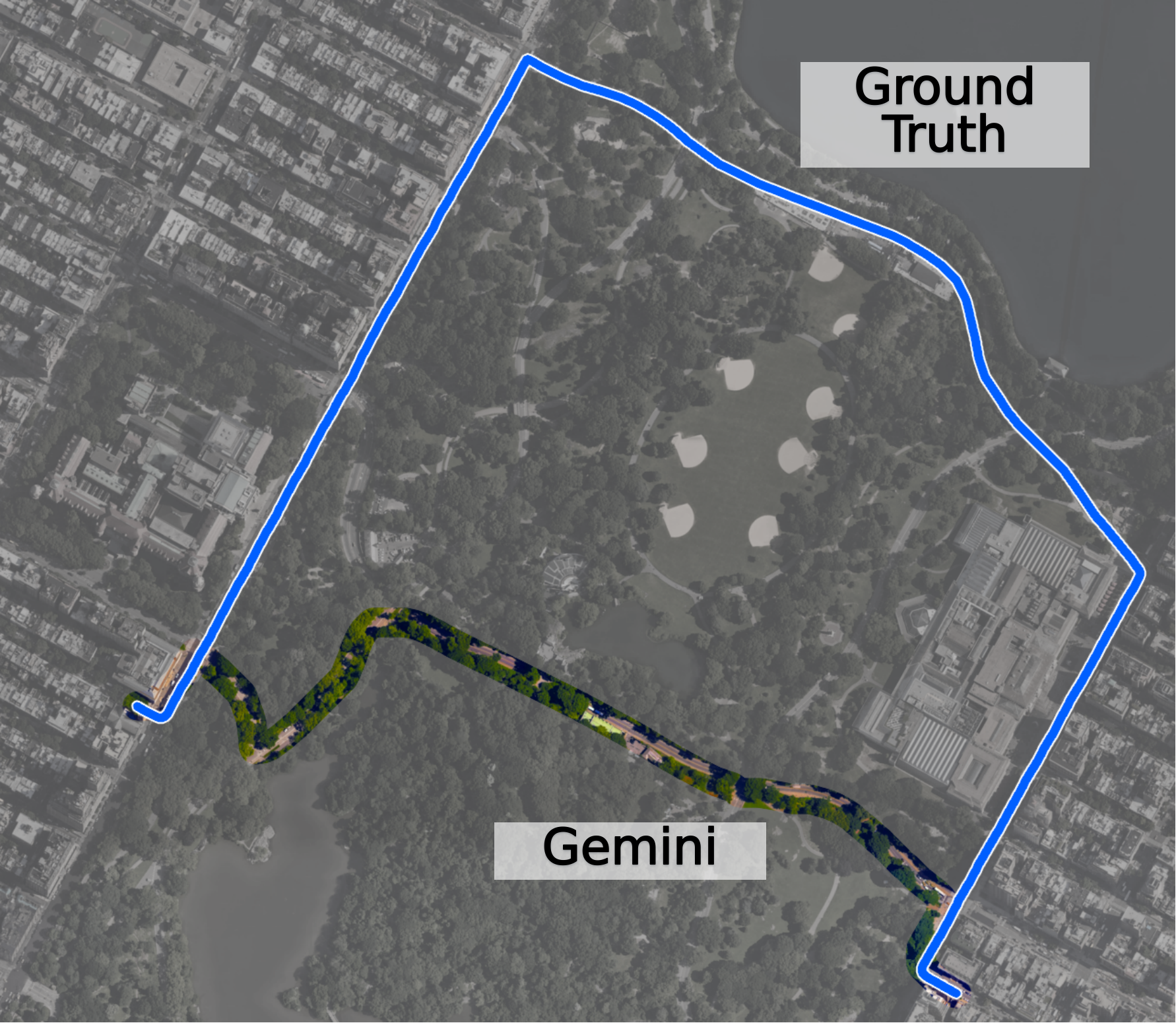}
    \vspace{-2mm} 
    \captionof{figure}{Gemini predicts roads unlabeled in the ground truth.}
    \label{fig:valid}
  \end{minipage}
\end{figure}


\paragraph{Downstream path-planning benchmark.}
To further evaluate the navigation utility of generated priors, we compare with road-topology extraction methods on a start-goal path-planning benchmark from CityScale~\cite{he2020sat2graph} and Global-Scale~\cite{yin2024satelliteimageroadgraph}. We also include SAM 3.1 as a segmentation baseline to test whether visible-road segmentation alone is sufficient, since planning requires continuous and connected traversable regions.
Both datasets provide high-resolution $2048 \times 2048$ aerial images with ground-truth road annotations. For each image, we randomly sample 1000 start-goal pairs and evaluate whether each method supports path planning between them. CityScale is used as the in-domain road-only benchmark, while the out-of-domain (OOD) split of Global-Scale evaluates cross-domain generalization. For fair comparison, all outputs are converted into binary road traversability masks and passed to the same $\mathrm{A}^*$ planner. Although our full system targets broader traversability, this benchmark is restricted to roads because existing public datasets provide road-only annotations. For Gemini, the prompt targets road traversability while encouraging global topological connectivity.

We report \emph{Succ.} as the fraction of sampled start-goal pairs where the planner finds a connected path on the predicted traversability mask. For successful cases, \emph{Valid.} measures the path fraction within the ground-truth traversable region, and \emph{Len.} denotes the predicted-to-ground-truth shortest-path length ratio, where values closer to 1 are preferred. Average inference time is also reported.

As shown in Table~\ref{tab:downstream_path_eval}, RNGDet++ and SAMRoad perform well on in-domain CityScale but degrade on Global-Scale OOD due to missing or disconnected road segments. SAM 3.1 also suffers from fragmented masks and missed narrow paths, making these baselines often impractical in real-world scenes (Fig.~\ref{fig:ooood}). In contrast, Gemini achieves higher OOD reachability and a more stable length ratio, showing stronger robustness for global route generation. Its lower validity is partly due to annotation mismatch, as Gemini may use visually plausible road-like regions that preserve connectivity but are unlabeled as roads (Fig.~\ref{fig:valid}). Overall, Gemini provides more reliable global path candidates than graph-based baselines with only a modest validity trade-off.


We also test Gemini-Direct, where the model directly generates a path-oriented mask between the start and goal, rather than a complete traversability prior. We then apply the same $\mathrm{A}^*$ planner to this path-oriented mask for fair comparison. Its lower success rate and unstable path quality
  show that direct path-mask generation is less reliable, often producing
  invalid
  shortcuts across non-traversable regions. Thus, our system uses the
  prior-generation-and-search pipeline.

%
%
%



\subsection{Real-world Results}
We deploy two types of near-ground platforms: a low-altitude quadrotor UAV and a quadruped robot. Both the quadrotor UAV and each quadruped robot are independently equipped with an Intel RealSense D455 camera for cross-view localization, a Livox Mid-360 LiDAR for state estimation and local mapping, and an NVIDIA Orin NX (16 GB) for onboard computation. To ensure safe near-ground flight, the UAV is constrained to a fixed altitude of 1\,m.
A prior BEV map is constructed from aerial images captured by a high-altitude DJI Neo drone and processed with WebODM into a georeferenced orthomosaic with scale information. For cross-view localization, we directly use the BEV-Patch-PF~\cite{lee2025bev} feature extractor without scene-specific fine-tuning. Due to onboard computation constraints, the cross-view localization module outputs global odometry at 1\,Hz, which is used only for periodic drift correction and local goal updates. The local planner runs independently at 10\,Hz using 200\,Hz odometry from FAST-LIO2~\cite{xu2022fast} fused with IMU measurements, enabling real-time obstacle avoidance and trajectory tracking on both platforms.
\begin{figure}[!h]
    \centering
    \includegraphics[width=1.0\linewidth]{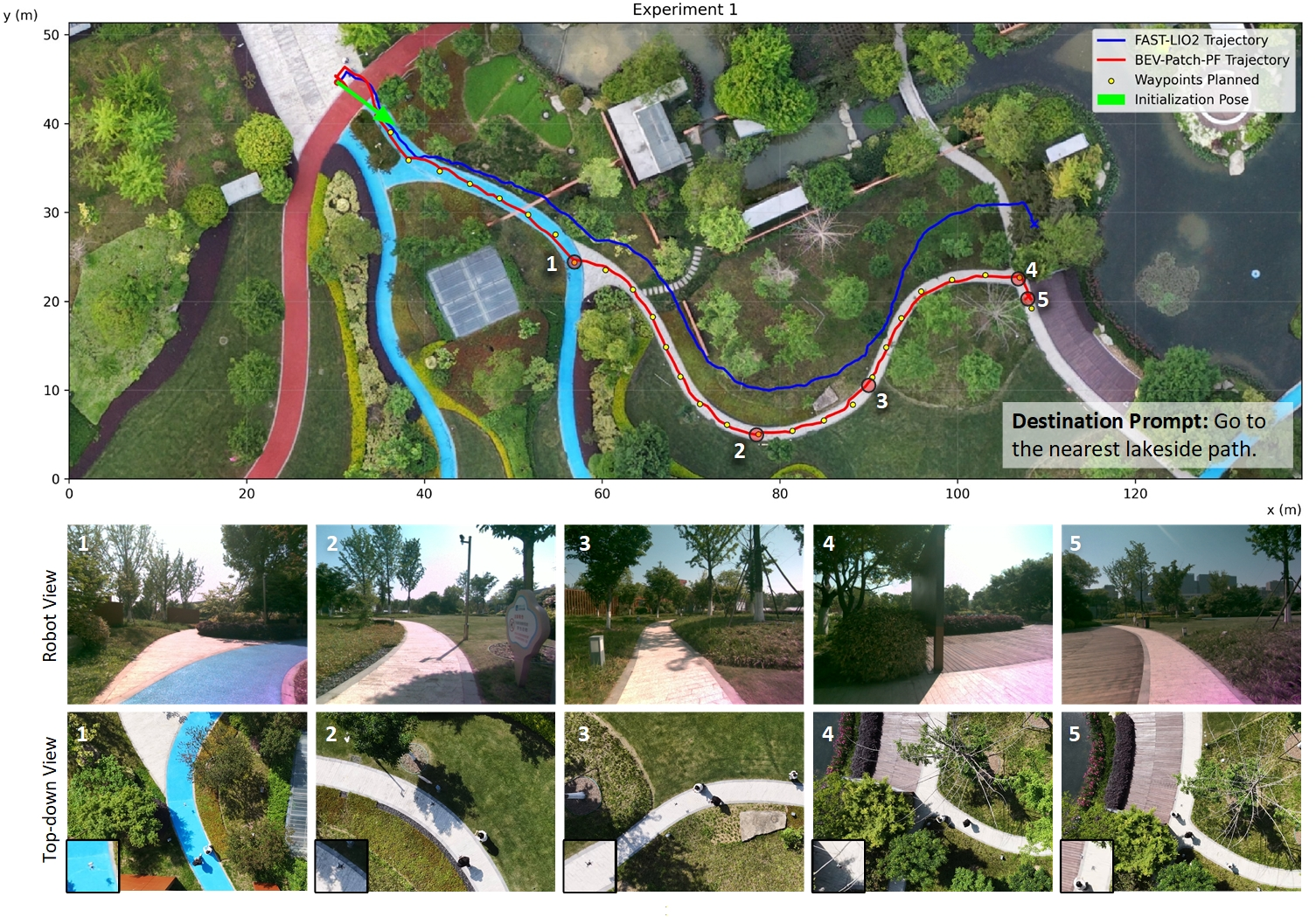}
    \caption{\textbf{Experiment 1.} Drone navigation in a park. The initial global pose estimate contains relatively large errors, making FAST-LIO2 alone insufficient for long-range navigation.}
    \label{fig:exp1}
\end{figure}

Since our local motion planner is lightweight and only handles low-level geometric obstacle avoidance and trajectory tracking, it does not reason about road-level semantics. 
As a result, small BEV-level path deviations may lead the robot away from the intended road region. 
To improve execution robustness, we adopt the Gemini method in Table~\ref{tab:downstream_path_eval} for Experiments 1 and 2, where we prompt the image-generation model to draw thin lines to reduce path ambiguity.
For Experiment 3, we use Gemini-Direct from Table~\ref{tab:downstream_path_eval}, which directly prompts Gemini to generate a navigation path. This experiment tests whether Gemini-Direct can correctly interpret the scene and produce a safe path without the full traversability-prior pipeline.


\begin{figure}[!t]
    \centering
    \includegraphics[width=1.0\linewidth]{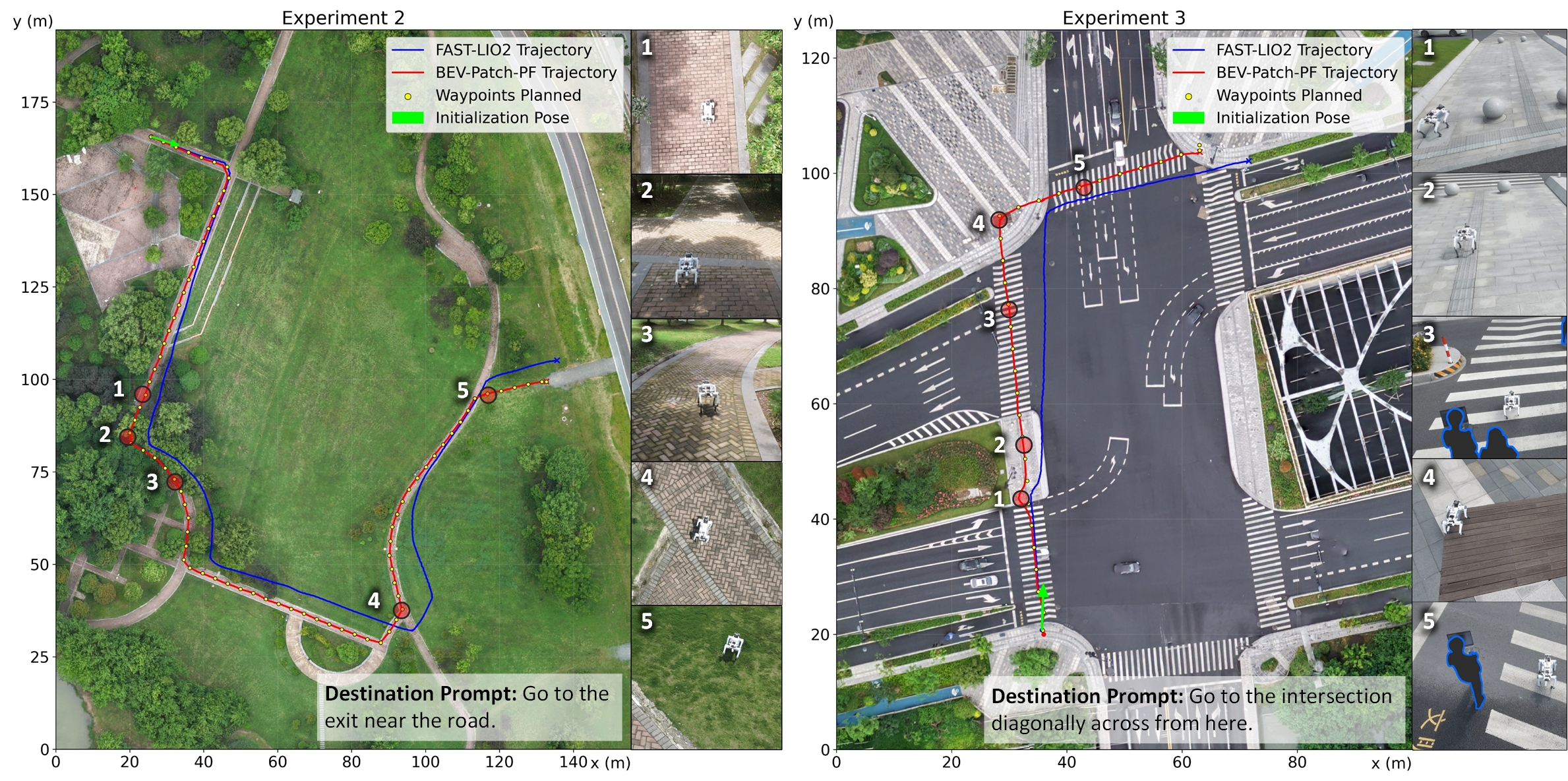}
    \caption{\textbf{Experiment 2:} Quadruped robot navigation in a park. \textbf{Experiment 3:} Quadruped robot navigation in an urban block.}
    \label{fig:exp2}
\end{figure}
\textbf{Experiment 1} (Fig.~\ref{fig:exp1}) is conducted in an urban park without annotated pathways in commercial mapping services, highlighting the need for online environmental understanding and path planning. Raw FAST-LIO2 exhibits large initial pose errors and long-term drift, making it unsuitable for direct long-range navigation. In contrast, the cross-view module provides continuous global localization and dynamic correction of the local navigation goal. Without reliable RTK/GPS ground truth, we qualitatively validate the trajectory using synchronized RGB and aerial footage with manually annotated waypoints, showing that the UAV follows the planned corridor.


\begin{figure}[!t]
    \centering
    \includegraphics[width=1\linewidth]{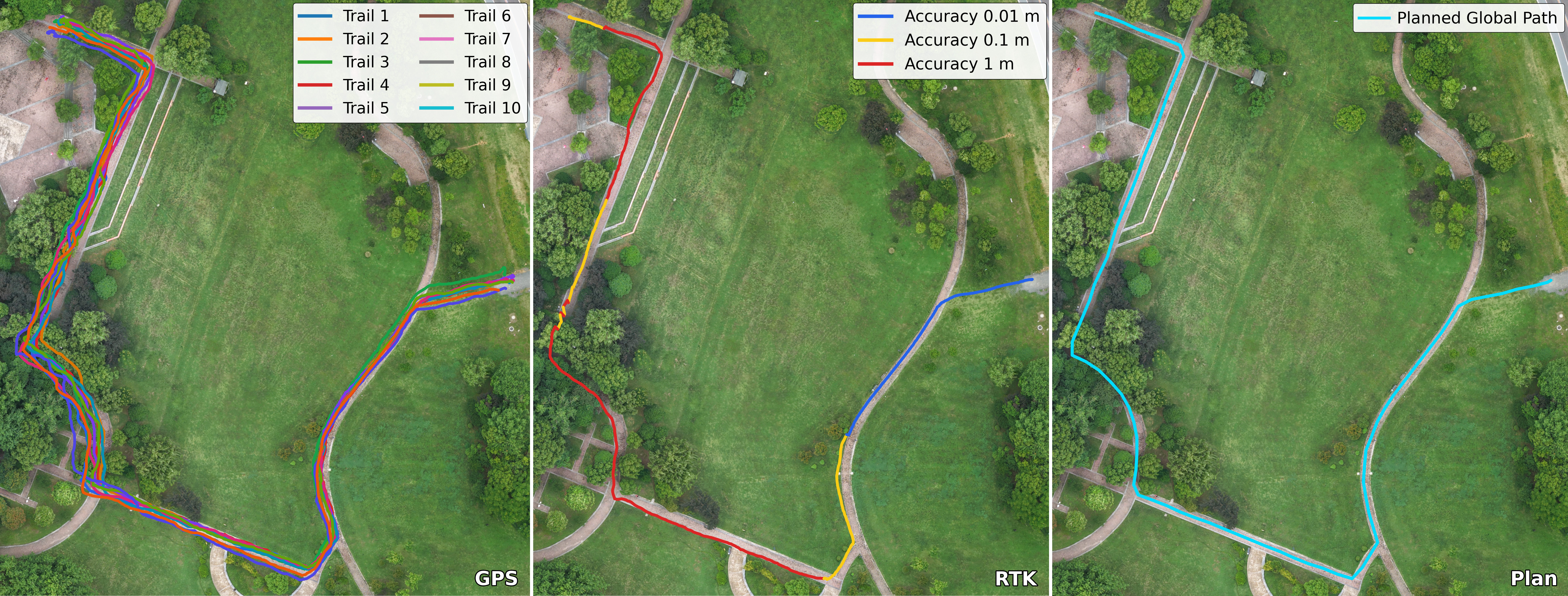}
    \caption{\textbf{Quantitative Real-World Navigation Results.} \textbf{Left:} GPS trajectories from 10 autonomous trials.\textbf{ Middle:} RTK trajectory from one trial, color-coded by positioning accuracy.\textbf{ Right: }Planned global path from the same trial.}
    \label{fig:placeholder}
\end{figure}

\textbf{Experiments 2 and 3} (Fig.~\ref{fig:exp2}) further highlight the strong generalization capability of the image-generation model in complex real-world environments. 
In Experiment 2, as shown around waypoints 2 and 3 in the left subfigure of Fig.~\ref{fig:exp2}, the road is severely occluded by dense tree canopies; nevertheless, the model infers the underlying traversable road and generates a feasible path. 
It also identifies a footpath formed by repeated pedestrian use and incorporates it into the planned trajectory. 
Experiment 3 is conducted at an urban intersection, where the planned trajectory reflects the model's awareness of traffic-related spatial semantics and road-use constraints. 

\textbf{Repeated Real-World Experiments}
We equipped a quadruped with an RTK/GPS receiver (WTRTK-982\footnote{\url{https://www.wit-motion.com/RTKPositioning/69.html}}) and conducted 10 consecutive autonomous trials along the 285~m route in Experiment~2. Nine trials succeeded without human intervention or collision (90\%); the sole failure resulted from localization error that diverted the robot from the intended goal. Figure~\ref{fig:placeholder} shows the GPS trajectories (left), RTK measurements (middle), and planned global path (right).

Despite occasional accuracy degradation under poor signal reception, RTK served as the best available reference. The mean tracking error relative to RTK was 0.901~m, with initial position and heading errors of 0.660~m and $4.360^\circ$. Mean cross-view localization, GPS, and LiDAR odometry errors were 0.852~m, 1.253~m, and 24.132~m, respectively.

Accurate RTK is not always available, GPS remains imprecise, and LiDAR odometry accumulates drift over long distances. These limitations motivate combining periodic cross-view global corrections with high-frequency local odometry for trajectory tracking.

\section{Limitations}
First, our lightweight local motion planner only performs geometric obstacle avoidance and trajectory tracking, and thus even small BEV-level path errors from imperfect annotations or cross-view localization drift may lead to noticeable execution deviations.
Future work could replace or augment the local planner with foundation-model-based navigation policies, such as route-conditioned VLA models~\cite{li2025urbanvla}. 
Second, the system relies on accurate georeferenced orthomosaics and reasonably accurate cross-view localization, which may limit deployment with outdated or poorly aligned aerial priors. 
This dependency could be reduced by converting PathPainter outputs into high-level guidance prompts, such as ``go straight'' or ``turn left''. 
Finally, image-generation models may hallucinate nonexistent or geometrically inconsistent paths, which could be mitigated by multi-sample consistency checks, traversability verification, and closed-loop replanning.

\section{Conclusion}

We propose a BEV-based navigation planning framework driven by an image generation model. By formulating both destination selection and traversability-mask generation as image generation tasks, and leveraging cross-view localization for reliable global odometry, our approach transfers the strong generalization capability of image generation models to long-range outdoor navigation.


\clearpage
\acknowledgments{This work was supported in part by the National Natural  
   Science Foundation of China under Grant No. 62322314, and in part by the    
   Fundamental Research Funds for the Central Universities under Project No.   
   226-2025-00114.} 



\bibliography{example}  
\newpage

\appendix
\section{Detailed Image Generation Process}
\label{app:image_generation}

\begin{figure}[htbp]
\centering
\includegraphics[width=0.95\textwidth]{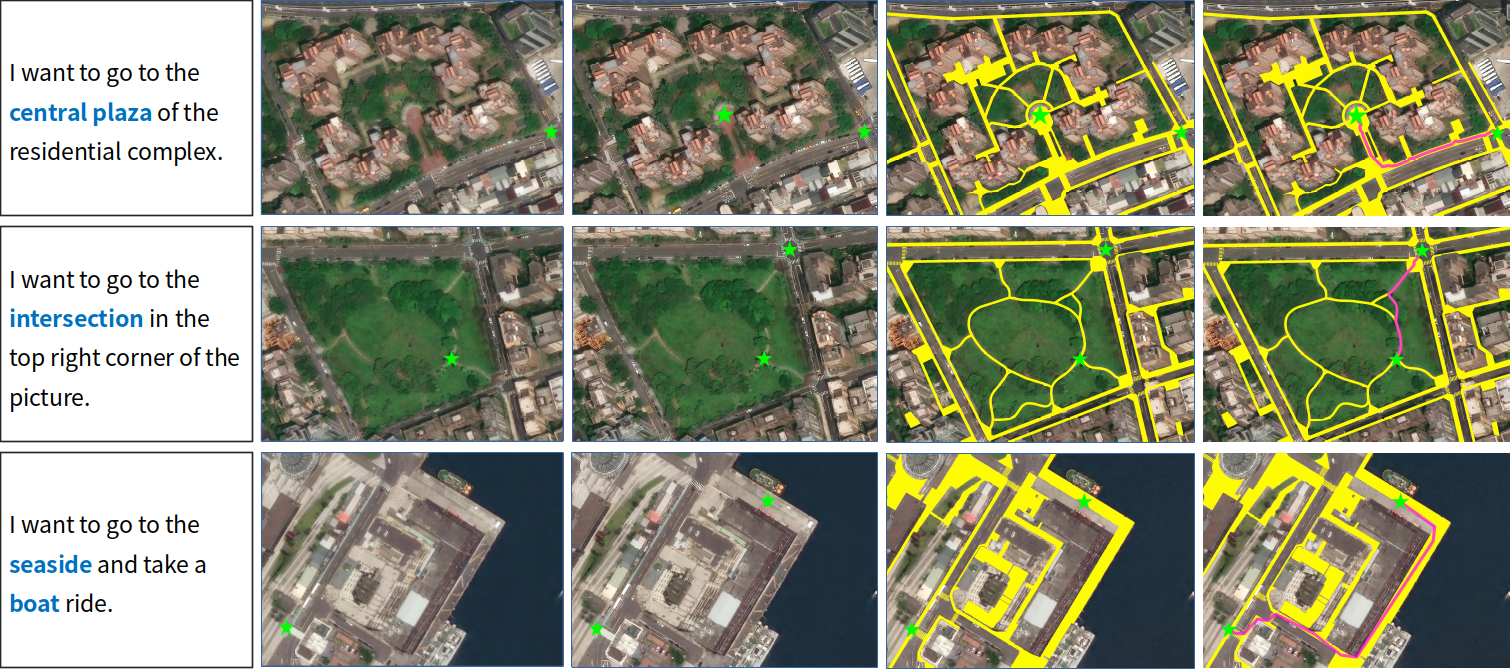}
\caption{\textbf{Complete pipeline visualization across three test scenarios. }The grid displays the continuous progression from the original image (with manually annotated start positions) through target goal localization (Step 1), pedestrian accessibility masking (Step 2), and the final optimal trajectory planned via the $\mathrm{A}^*$ algorithm (Step 3).}
\label{fig:pipeline_appendix}
\end{figure}

This section details the pipeline for goal localization, pedestrian accessibility mapping, and path planning using aerial imagery. The complete visual workflow across three diverse scenarios is illustrated in Figure~\ref{fig:pipeline_appendix}, followed by the exact prompts utilized for each computational step.

\subsection{LLM Prompts for Target Navigation}

\subsubsection{Step 1: Goal Localization Prompt}
The LLM is provided with the following template to mark the target coordinate based on sparse text instructions. The third sentence is dynamically updated based on the specific user task:

\begin{tcolorbox}[
  colback=gray!8,
  colframe=gray!50,
  arc=4pt,
  boxrule=0.6pt,
  title={\small\textbf{Step 1: Goal Localization Prompt}},
  fonttitle=\bfseries,
  left=6pt, right=6pt, top=4pt, bottom=4pt
]
\small\ttfamily
You are an expert in analyzing aerial imagery to determine pedestrian accessibility.\\
The input data is an aerial view of an urban area.\\
{}[Dynamic Destination Instruction, e.g., ``I want to go to the seaside and take a boat ride.'']\\
Please mark it with a solid bright green (\#00FF00) pentagram in the original image. 
\end{tcolorbox}

\subsubsection{Step 2: Pedestrian Accessibility Masking Prompt}
To build the costmap for path planning, the LLM generates a pixel-level dense pedestrian mask using the comprehensive structured instructions below:

\begin{tcolorbox}[
  breakable,
  colback=gray!8,
  colframe=gray!50,
  arc=4pt,
  boxrule=0.6pt,
  title={\small\textbf{Step 2: Pedestrian Accessibility Masking Prompt (Gemini)}},
  fonttitle=\bfseries,
  left=6pt, right=6pt, top=4pt, bottom=4pt
]
\small\ttfamily
You are an expert in analyzing aerial imagery to map pedestrian accessibility.\\
Input: An aerial/satellite image of an urban area.\\
Objective: Identify and annotate all walkable zones to produce a complete pedestrian accessibility map.\\
What to mark as walkable (highlight in bright yellow \#FFFF00):
\begin{itemize}[noitemsep,topsep=0pt,leftmargin=15pt]
    \item Sidewalks, footpaths, and pedestrian lanes,
    \item Crosswalks and zebra crossings,
    \item Plazas, courtyards, and open pedestrian areas,
    \item Pedestrian bridges, underpasses, and elevated walkways,
    \item Any other surface reasonably accessible on foot,
    \item Paths beneath tree canopy --- if a walkable surface is visible or inferable under trees, mark it as walkable.
\end{itemize}
What to exclude (do not mark):
\begin{itemize}[noitemsep,topsep=0pt,leftmargin=15pt]
    \item Building interiors, rooftops,
    \item Walls, fences, and barriers,
    \item Waterways, water bodies,
    \item Roads, lanes, and vehicle-only areas,
    \item Restricted or clearly inaccessible zones.
\end{itemize}
General connectivity rule: For other discrete walkable segments (e.g., a sidewalk interrupted by a small gap), bridge them into a single contiguous region to avoid fragmentation.\\
Crosswalk/Zebra crossing fill rule: Always render each crosswalk as a solid filled rectangle (bounding box), not as individual stripes or parallel lines.\\
Preservation rule: Do not remove, overwrite, or alter any existing markers (e.g., star icons) already present on the image. Overlay the yellow annotation on top of the original image without modifying other visual elements.\\
Coverage rule: Systematically scan the entire image from edge to edge. Do not skip peripheral or partially visible areas. Every walkable segment must be annotated to ensure full coverage.\\
Output: The original image with all walkable areas filled or outlined in bright yellow (\#FFFF00), forming a complete, connected pedestrian accessibility map.
\end{tcolorbox}

\begin{figure}[htbp]
\centering
\includegraphics[width=0.95\textwidth]{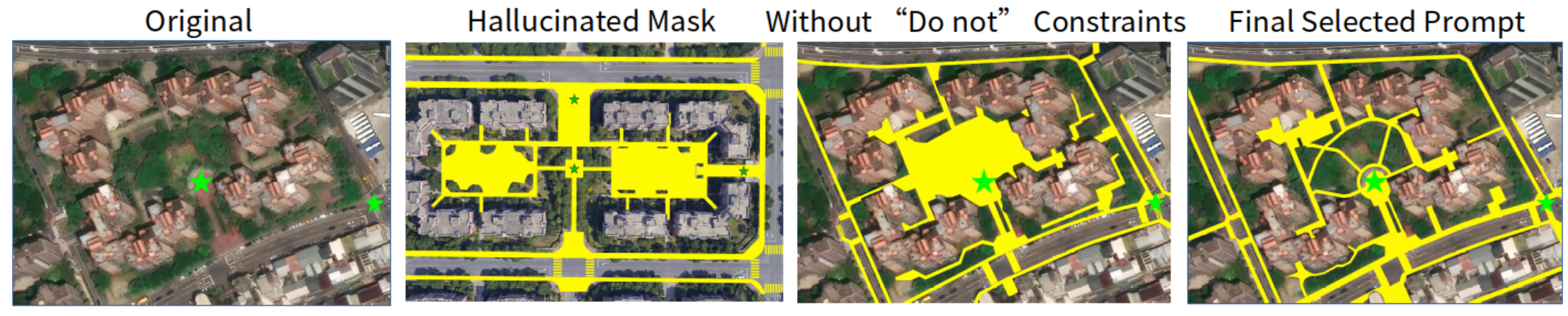}
\caption{
\textbf{Prompt constraint trade-off analysis for LLM-generated pedestrian accessibility masks. }From left to right:
(1) Original aerial image.
(2) Using the full restrictive prompt may still produce occasional hallucinations.
(3) Removing the restrictive ``Do not'' constraints significantly degrades mask quality, causing severe over-segmentation into vehicle lanes and forests.
(4) Final accessibility mask generated using the selected prompt configuration, which achieves the best practical balance between hallucination suppression and segmentation quality.
}

\label{fig:mask_tradeoff}
\end{figure}

In practice, we observe that the prompt design introduces an inherent trade-off between hallucination suppression and segmentation quality. As illustrated in Fig.~\ref{fig:mask_tradeoff}, the large number of restrictive ``Do not'' constraints may still occasionally produce hallucinated pedestrian discontinuities or missing walkable connections. However, removing these restrictive constraints significantly degrades the generated accessibility masks, often causing severe over-segmentation into vehicle lanes, roads, and building regions.

Therefore, we retain the restrictive prompt structure as a practical deployment trade-off. The downstream $\mathrm{A}^*$ planner serves as a consistency check: if no feasible path is found, the system concurrently regenerates five accessibility masks and adopts the first one that succeeds. If all attempts fail, the destination is reported unreachable under the current environmental understanding.


\section{Benchmark Examples}
\subsection{Visible-road Segmentation benchmark}
This section presents representative examples selected from the visible-road segmentation benchmark in Table~\ref{tab:path_generation_eval}. 
For GPT and Gemini, we use the same complete prompt:

\begin{tcolorbox}[
  colback=gray!8,
  colframe=gray!50,
  arc=4pt,
  boxrule=0.6pt,
  title={\small\textbf{Visible-road Segmentation Benchmark Prompt}},
  fonttitle=\bfseries,
  left=6pt, right=6pt, top=4pt, bottom=4pt
]
\small\ttfamily
Generate a semantic segmentation visualization image, using this color 
mapping: {"road": <255, 255, 0>, "background": <0, 0, 0>}.
\end{tcolorbox}

\begin{figure}[h]
    \centering
    \includegraphics[width=1.0\linewidth]{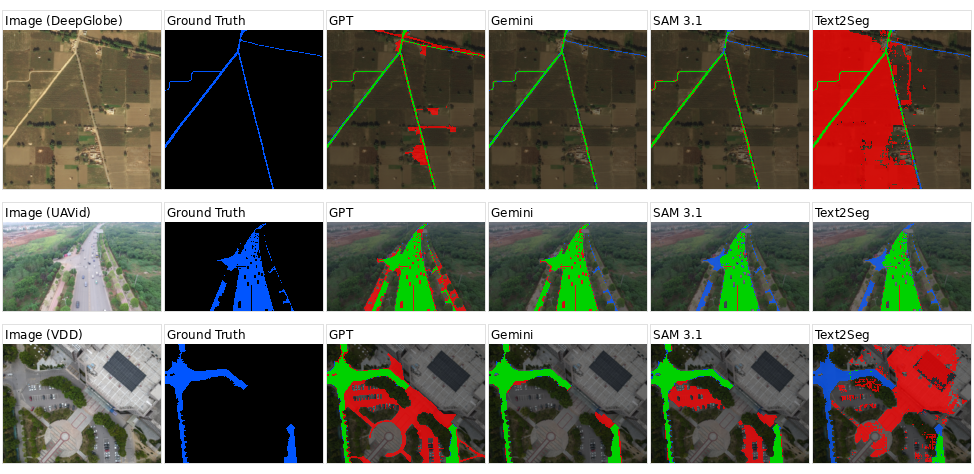}
    \caption{\textbf{Example results from the Visible-road Segmentation benchmark.} In the comparison images for each example, green indicates correctly predicted regions, red indicates incorrectly predicted regions, and blue indicates missed regions.
}
    \label{fig:weqwe}
\end{figure}

In our experiments, as shown in Fig \ref{fig:weqwe}, generative models often predicted on-road vehicles and locally occluded regions as road areas. This suggests that their outputs are not strict visible-pixel segmentations, but rather road-region estimates with contextual semantic completion. In visible-road evaluation, this behavior leads to false positives and reduced precision, while potentially helping preserve road connectivity.

In addition, we found that GPT was more prone to hallucinations than Gemini under complex, lengthy text prompts, and that Gemini outperformed GPT on the visible-road segmentation benchmark. Therefore, we ultimately selected Gemini as the generative model for all subsequent tasks.

\subsection{Downstream Path-planning Benchmark}
For Gemini, we use the prompt below:

\begin{tcolorbox}[
  breakable,
  colback=gray!8,
  colframe=gray!50,
  arc=4pt,
  boxrule=0.6pt,
  title={\small\textbf{Downstream Path-planning Benchmark Prompt (Gemini)}},
  fonttitle=\bfseries,
  left=6pt, right=6pt, top=4pt, bottom=4pt
]
\small\ttfamily
Generate a topology-complete road-network centerline map for the input aerial image.\\
You are an expert in aerial road-network reasoning and amodal road completion. Treat the image as a top-down satellite view. Your task is not only to trace visible roads, but to infer the most plausible continuous drivable road network when the road is temporarily hidden.\\
Draw the final road network as pure green (0, 255, 0) centerlines on a pure black (0, 0, 0) background. Use only these two colors. The green centerlines should be about 2-3 pixels wide, fully opaque, smooth, solid, and easy to extract.\\
 Dream the missing topology where the evidence is strong. If a road is briefly hidden by tree canopy, shadows, vehicles, bridge decks, overpasses, signs, lane markings, low contrast, or small image artifacts, complete the road through the hidden part. Continue the centerline smoothly using the visible road direction, width, alignment, curvature, and nearby connected road geometry.\\
Preserve connectivity as the primary goal. Roads that are clearly the same logical road should be one continuous connected stroke, not separated into fragments. Intersections, branches, ramps, bridge approaches, service roads, and road continuations must physically touch with no tiny gaps. Prefer a short plausible connector over a broken line when both sides strongly align.\\
Be imaginative only for short, local, visually supported gaps. Do not invent long new roads.
\end{tcolorbox}

For Gemini-Direct, we use the prompt below:

\begin{tcolorbox}[
breakable,
colback=gray!8,
colframe=gray!50,
arc=4pt,
boxrule=0.6pt,
title={\small\textbf{Downstream Path-planning Benchmark Prompt (Gemini-Direct)}},
fonttitle=\bfseries,
left=6pt, right=6pt, top=4pt, bottom=4pt
]
\small\ttfamily
You are an expert pedestrian route planner analyzing a top-down aerial image of an urban area. The image has two bright magenta (\#FF00FF) five-point stars marking START and END. Draw one continuous single walkable path from the center of START to the center of END on top of the aerial image. Use pedestrian-friendly surfaces whenever possible: sidewalks, crosswalks, pedestrian plazas and bridges. Avoid building interiors, walls, fences, water, and highways. Tree canopy does not block walkability. Draw the path as a single bright magenta (\#FF00FF) line, 4 pixels wide, roughly following the underlying walkway centerlines; the line must be unbroken from START to END. Do not move or alter the START and END markers, and do not add any other annotations or colors.You are an expert pedestrian route planner analyzing a top-down aerial image of an urban area. The image has two bright magenta (\#FF00FF) five-point stars marking START and END. Draw one continuous single walkable path from the center of START to the center of END on top of the aerial image. Use pedestrian-friendly surfaces whenever possible: sidewalks, crosswalks, pedestrian plazas and bridges. Avoid building interiors, walls, fences, water, and highways. Tree canopy does not block walkability. Draw the path as a single bright magenta (\#FF00FF) line, 4 pixels wide, roughly following the underlying walkway centerlines; the line must be unbroken from START to END. Do not move or alter the START and END markers, and do not add any other annotations or colors.
\end{tcolorbox}

\begin{figure}[!t]
    \centering
    \includegraphics[width=1\linewidth]{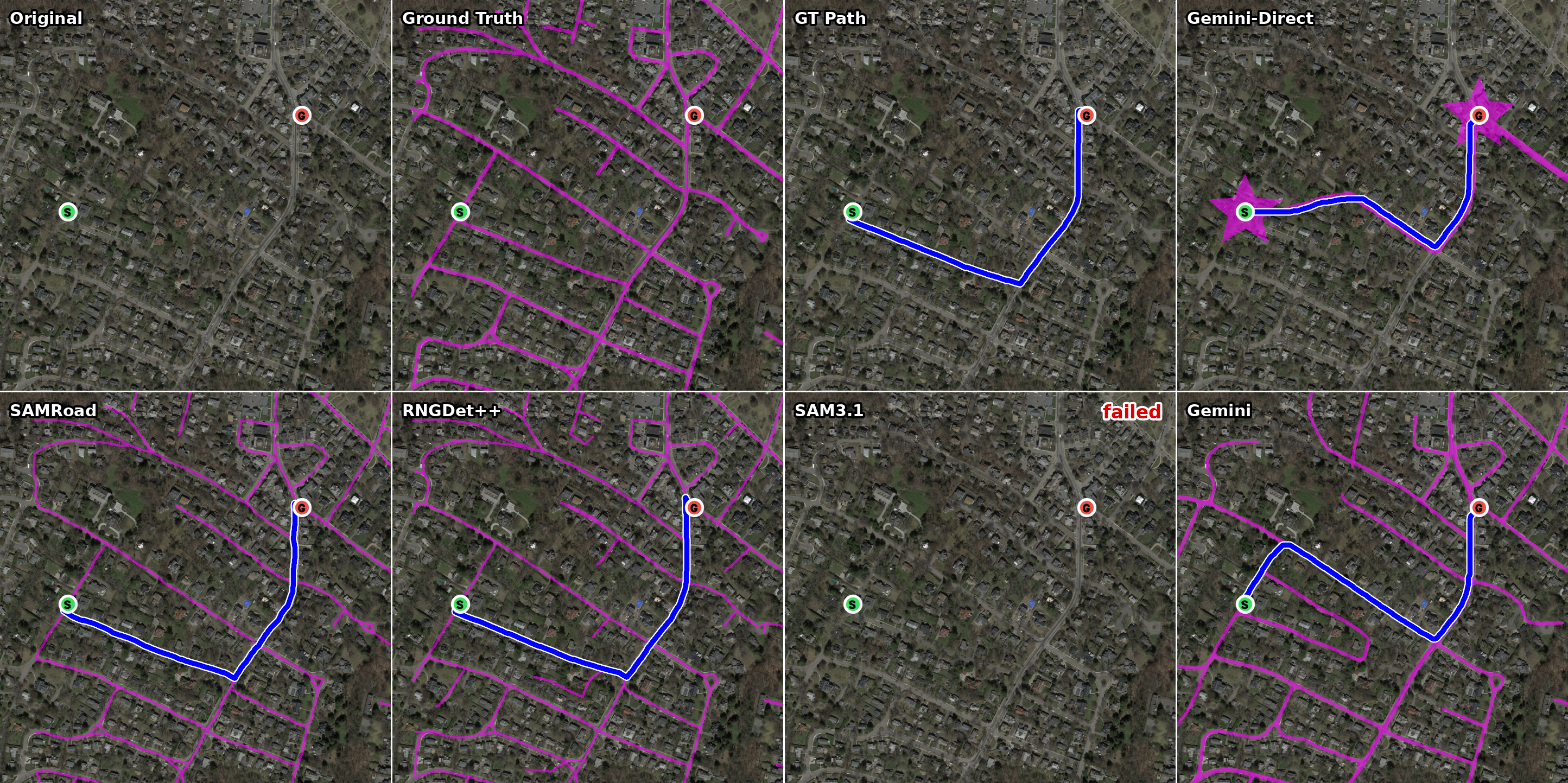}
    \caption{\textbf{In-domain(CityScale) sample.} The Gemini-Direct method generated an invalid path that does not exist at all, cutting diagonally through the block, while SAM 3.1 failed to segment the road area in the image.}
    \label{fig:indomain1}
\end{figure}

\begin{figure}[!h]
    \centering
    \includegraphics[width=1\linewidth]{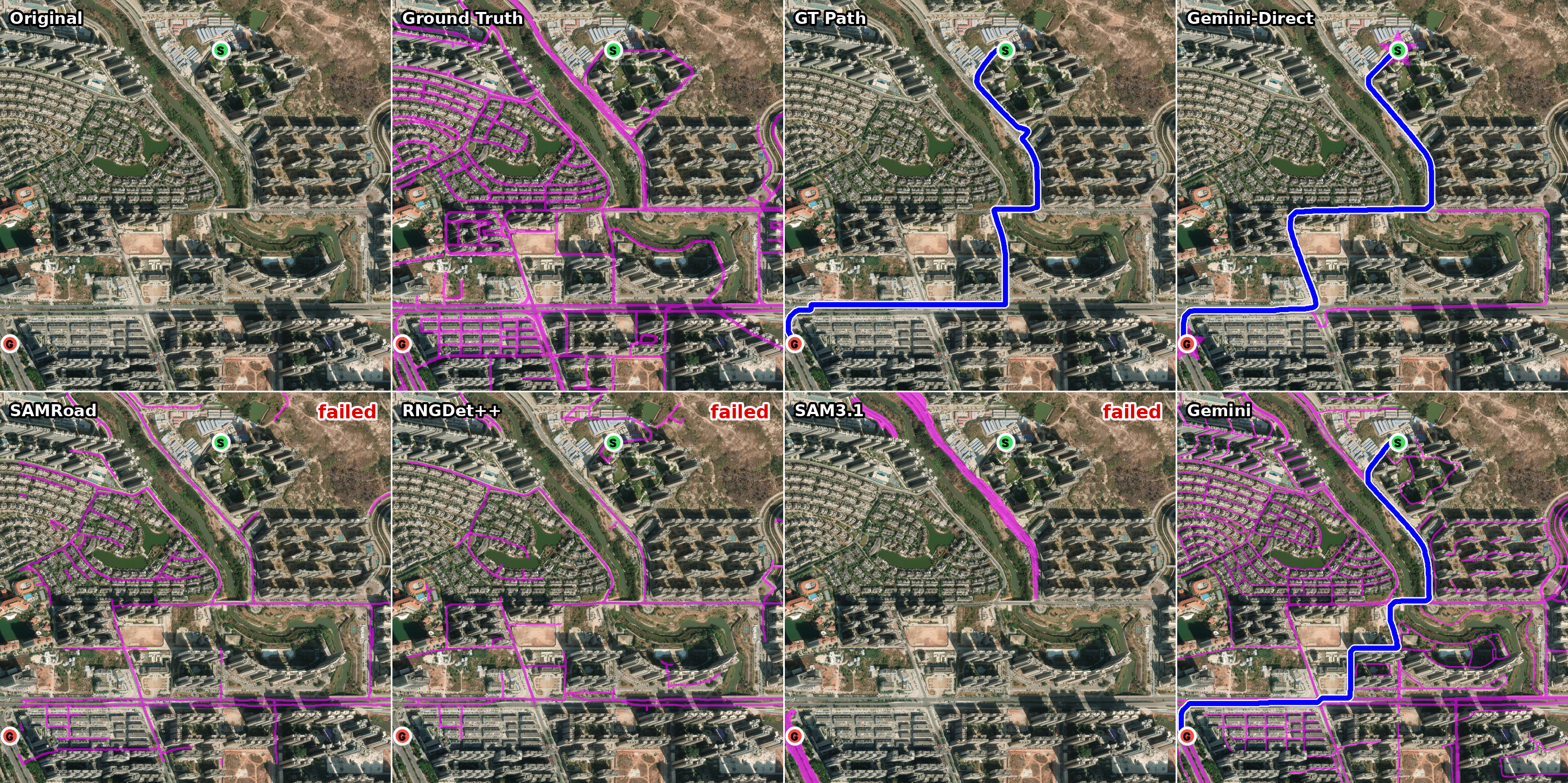}
    \caption{\textbf{OOD(Global-Scale) sample. }SAMRoad and RNGD++ perform poorly on OOD data, often failing to generate continuous and complete road topology.}
    \label{fig:ood1}
\end{figure}

We provide two representative qualitative examples from the in-domain and OOD data, respectively. As shown in Fig. \ref{fig:indomain1} and Fig. \ref{fig:ood1}, each example includes the original dataset image, the ground-truth path, and the start and goal points used for testing, denoted by S and G, respectively. For each method, the extracted road region is visualized with a magenta mask, while the planned shortest path is shown as a blue solid line. If a method fails to generate any valid path, we mark it as ``failed'' in the upper-right corner.

As shown in Fig. \ref{fig:indomain1}, Gemini-Direct can sometimes produce invalid paths that directly cross non-traversable regions, indicating unstable behavior when the model is asked to generate the path directly. In contrast, the Gemini-based method produces a coherent road mask for the urban block, providing a more reliable basis for subsequent shortest-path planning.

However, when applied to OOD data, as shown in Fig. \ref{fig:ood1}, although the input image still depicts an urban block, road extraction methods trained on specific datasets often fail to recover large-scale continuous road topology under distribution shift. This fragmentation prevents valid path generation and leads to a significant drop in success rate. By contrast, the generative method benefits from the strong generalization capability of foundation models, allowing it to maintain more stable performance across both in-domain and OOD scenarios.

\section{More Details on Real-world Experiments}
\label{realworld}

\begin{figure}[!h]
    \centering
    \includegraphics[width=1\linewidth]{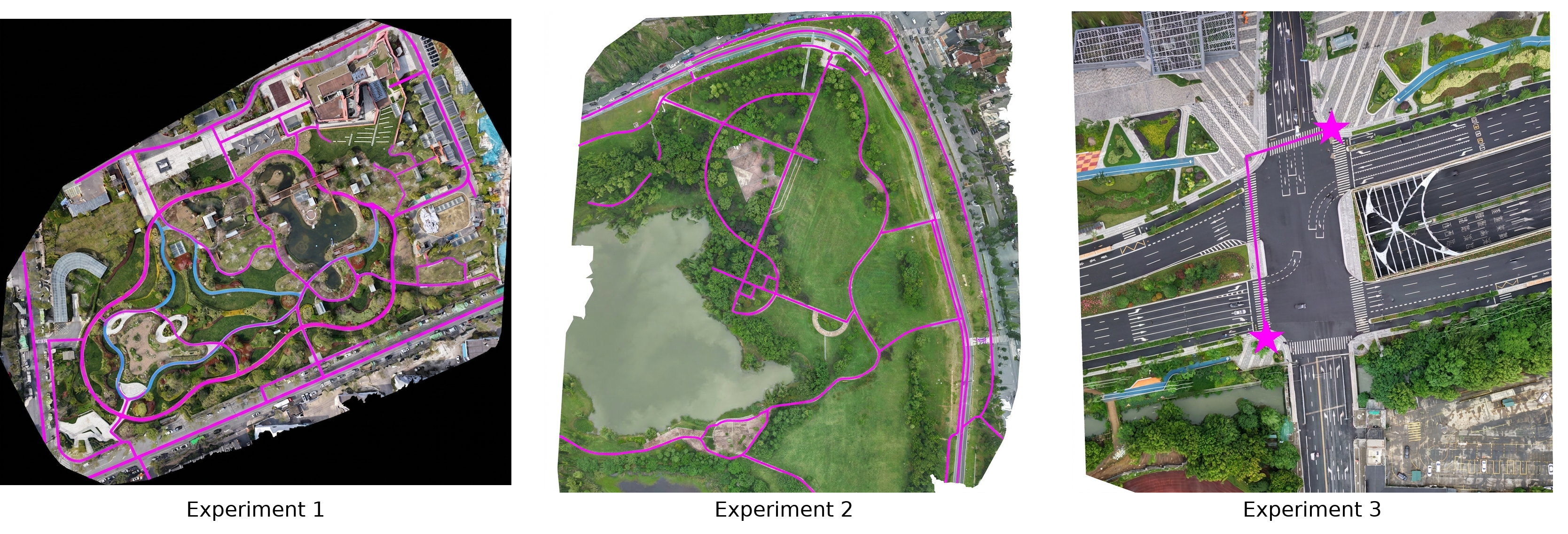}
    \caption{Generated images used in real-world experiments.}
    \label{fig:generatedimagesusedinrealworld}
\end{figure}

The generated images used in the real-world experiments are shown in Fig.~\ref{fig:generatedimagesusedinrealworld}. Since our local motion planner lacks semantic understanding, the extracted path must be aligned as closely as possible with the road centerline. To this end, we prompt the image-generation model to render roads or road topology with a thin line width of 2--3 pixels.

\end{document}